\documentclass[pdflatex,sn-mathphys-num]{sn-jnl}

\usepackage{graphicx}%
\usepackage{multirow}%
\usepackage{amsmath,amssymb,amsfonts}%
\usepackage{amsthm}%
\usepackage{mathrsfs}%
\usepackage[title]{appendix}%
\usepackage{xcolor}%
\usepackage{textcomp}%
\usepackage{manyfoot}%
\usepackage{booktabs}%
\usepackage{algorithm}%
\usepackage{algorithmicx}%
\usepackage{algpseudocode}%
\usepackage{listings}%
\usepackage{url}
\usepackage{hyperref}
\usepackage{siunitx}

\theoremstyle{thmstyleone}%

\theoremstyle{thmstyletwo}%

\theoremstyle{thmstylethree}%

\begin{document}

\title[Robust Exploration in DCS via RL with Soft MoE]{Robust Exploration in Directed Controller Synthesis via Reinforcement Learning with Soft Mixture-of-Experts}

\author[1]{\fnm{Toshihide} \sur{Ubukata}}
\author[2]{\fnm{Zhiyao} \sur{Wang}}
\author[3]{\fnm{Enhong} \sur{Mu}}
\author*[1,4]{\fnm{Jialong} \sur{Li}}\email{lijialong@fuji.waseda.jp}
\author[4]{\fnm{Kenji} \sur{Tei}}

\affil[1]{\orgname{Waseda University}, \orgaddress{\city{Tokyo}, \postcode{169-8050}, \country{Japan}}}
\affil[2]{\orgname{Osaka University}, \orgaddress{\country{Japan}}}
\affil[3]{\orgname{Southwest University}, \orgaddress{\country{China}}}
\affil[4]{\orgname{Institute of Science Tokyo}, \orgaddress{\city{Tokyo}, \postcode{152-8550}, \country{Japan}}}

\abstract{On-the-fly Directed Controller Synthesis (OTF-DCS) mitigates state-space explosion by incrementally exploring the system and relies critically on an exploration policy to guide search efficiently. Recent reinforcement learning (RL) approaches learn such policies and achieve promising zero-shot generalization from small training instances to larger unseen ones. However, a fundamental limitation is anisotropic generalization, where an RL policy exhibits strong performance only in a specific region of the domain-parameter space while remaining fragile elsewhere due to training stochasticity and trajectory-dependent bias.
To address this, we propose a Soft Mixture-of-Experts framework that combines multiple RL experts via a prior-confidence gating mechanism and treats these anisotropic behaviors as complementary specializations.
The evaluation on the Air Traffic benchmark shows that Soft-MoE substantially expands the solvable parameter space and improves robustness compared to any single expert.}


\maketitle

\section{Introduction}
\label{sec:introduction}

Controller synthesis is a central problem in formal methods, aiming to automatically generate a control strategy (or ``controller'') for a given system specification \cite{10.1145/3643915.3644095, 10174043}. The goal is to ensure that the system, while interacting with an uncertain environment, adheres \cite{Wonham} to a set of critical properties, which typically include safety, or non-blocking, which ensures the system can always eventually reach a desired goal state. In the context of Discrete Event Systems (DES), the system is often modeled as a parallel composition of interacting finite automata. However, when these automata are composed, the size of their global state space grows exponentially with the number of components. This phenomenon, known as the `state-space explosion', poses a major computational barrier for traditional monolithic synthesis approaches \cite{Cassandras:2006:DES}.

To counter the challenge of state-space explosion, the method of On-the-fly Directed Controller Synthesis (OTF-DCS) has been proposed \cite{10.1007/3-540-48119-2_15}. Drawing inspiration from heuristic search in automated planning, it reframes the synthesis problem as a search task within a vast, implicitly defined state space. Unlike monolithic methods that construct the full state space upfront, OTF-DCS maintains a partial view of the explored space, which is expanded incrementally at its `exploration frontier'. At each step, the algorithm selects a transition to expand from the frontier based on an exploration policy, adds it to the partial view, and updates its assessment of system states (e.g., classifying them as `winning' or `losing'). This process shifts the primary computational bottleneck from memory consumption to search efficiency, providing a scalable foundation for runtime self-adaptation \cite{10.1145/2897053.2903381} and the assured dynamic update of controllers when environment assumptions or requirements change \cite{10.1145/2897053.2897056}.

To enhance the efficacy of OTF-DCS, heuristic-based informed search strategy -- Ready Abstraction -- has been proposed, which analyzes the system's modular structure to estimate the distance to a goal state \cite{CiolekDCS}. The most recent paradigm involves using Reinforcement Learning (RL) to automatically learn a domain-specific exploration policy, to minimize the number of exploration steps \cite{Delgado_Sanchez}. The main advantage of the RL approach lies in its zero-shot generalization capability: a policy
\footnote{In this paper, the ``policy'' denotes the RL-based exploration heuristic used to guide the search process during synthesis, while the ``controller'' refers to the formal supervisory strategy generated as the final output of DCS. For detailed definitions, please refer to \cite{Delgado_Sanchez}.}
trained on small-scale instances can be directly applied to solve larger, more complex instances within the same problem domain, thus opening new possibilities for addressing large-scale synthesis problems.

However, while the RL approach has already demonstrated state-of-the-art performance, we acknowledge that it faces a fundamental challenge, namely anisotropic generalization. This phenomenon refers to the uneven distribution of a policy's performance across different environmental conditions, where high success rates concentrate in certain scenarios, while other regions remain fragile and error-prone. This issue is intrinsic to RL, as a policy's behavior is fundamentally shaped by its unique exploration trajectory during training. As a result, even policies trained under identical environments and hyperparameters can exhibit markedly different capabilities due to inherent stochasticity. Previous research has implicitly mitigated this intra-setting variance through a selection-based philosophy: train multiple models and deploy only the best-performing one \cite{Delgado_Sanchez}. However, we contend that this selection-based approach merely alleviates the symptoms of anisotropic generalization without addressing the root cause. While the selected champion may exhibit relatively superior overall performance, it is still a single policy constrained by its unique training history. In other words, its performance map remains inherently anisotropic; the selection mechanism simply wagers on finding a policy whose fragile regions are less problematic than those of others.

To this end, we propose a Mixture-of-Experts (MoE) framework that leverages the diversity of anisotropic specializations across different policies to learn a single, robust exploration policy.
The core tenet of this philosophy is to treat the anisotropic nature of individual policies not as a defect to be filtered out, but as a source of specialized, complementary expertise. It posits that the adaptability and robustness emerge not from identifying a single, universally strong policy, but from the synergistic integration of multiple policies, each mastering a different niche of the conditions.
The initial idea for this work was first presented in \cite{SEAMS2026}. This extended version introduces two major improvements: First, we refined the algorithm by implementing a prior-confidence gating mechanism that utilizes Gaussian prior interpolation and uncertainty metrics like entropy and margin to enable robust soft mixing. Second, we expanded the evaluation to include fine-grained efficiency metrics, such as the specific number of steps required for synthesis, and analyzed the expert selection strategies within the experiments.

The contributions of this paper are as follows:
\begin{itemize}
    \item We propose a Soft MoE framework that employs a prior-confidence gating mechanism to mitigate the anisotropic generalization issue in directed controller synthesis.
    \item We provide a detailed evaluation on an air traffic benchmark, demonstrating that our framework significantly expands the solvable parameter space compared to single experts.
\end{itemize}

The rest of this paper is organized as follows.
Section~\ref{sec:background} provides the necessary background on directed controller synthesis and the relevant RL-based approach.
Section~\ref{sec:proposal} details our proposed MoE framework for learning a robust policy.
Section~\ref{sec:evaluation} presents the evaluation of our approach, assessing its effectiveness and analyzing the contribution of its key components.
Finally, Section~\ref{sec:related} introduces related work and
Section~\ref{sec:conclusion} concludes with a summary and directions for future research.

\section{Background}
\label{sec:background}

This section provides the theoretical foundations for our approach. We begin by formalizing DCS and the specific optimization problem inherent in on-the-fly synthesis. We then discuss the application of RL to this problem and the challenge of anisotropic generalization, followed by the fundamentals of MoE.

\subsection{On-the-fly Directed Controller Synthesis}
\label{sec:background_dcs}

\subsubsection{Discrete Event Systems and Synthesis}
We adopt the standard framework of Supervisory Control Theory (SCT)~\cite{cassandras2008introduction}. A Discrete Event System (DES) is modeled as a deterministic finite automaton $E = (S, A, \delta, s_0, M)$, where $S$ is a finite set of states, $A$ is a finite set of event labels partitioned into controllable ($A_c$) and uncontrollable ($A_u$) events, $\delta: S \times A \to S$ is a partial transition function, $s_0 \in S$ is the initial state, and $M \subseteq S$ is the set of marked (goal) states.

In modular synthesis, the system is typically defined as the parallel composition of multiple components $E = E_1 || \dots || E_n$. The goal of controller synthesis is to generate a function $\sigma: A^* \to 2^{A_c}$ (the controller) that dynamically disables controllable events to ensure the system satisfies specified properties, typically safety (avoiding bad states) and non-blocking (always reaching $M$).

\subsubsection{Exploration Optimization Problem}
The monolithic construction of the composed plant $E$ suffers from state-space explosion. To mitigate this, On-the-fly Directed Controller Synthesis (OTF-DCS)~\cite{CiolekDCS, 8876664} incrementally explores the state space starting from $s_0$.
The exploration state at any step is defined by the \textit{Exploration Frontier} $F(E, h)$, which is the set of all unexplored transitions originating from states discovered in the current exploration history $h$.

The efficiency of OTF-DCS depends critically on the order in which transitions from $F(E, h)$ are expanded. This can be formulated as an optimization problem: finding an exploration policy $\pi$ that selects the next transition $a \in F(E, h)$ to minimize the total number of expansions $|h|$ required to synthesize a valid controller.
An effective policy acts as a heuristic, guiding the search toward the goal and pruning irrelevant paths, while a poor policy leads to state-space explosion or dead-ends.

\subsection{RL-based Exploration Policy}
\label{sec:background_rl}

Recent work has framed this exploration optimization as a Reinforcement Learning (RL) problem~\cite{Delgado_Sanchez} to learn a policy that estimates the value of expanding a specific transition. Under this formulation, the synthesis process is modeled as a Markov Decision Process (MDP) where the state represents the current partially explored graph (history), the action consists of selecting a transition to expand from the current frontier, and the reward is defined as a constant negative value -1 per expansion to encourage the agent to find the shortest path to a solution.

Training RL agents directly on large-scale DCS instances is computationally prohibitive due to the exponential growth of the state space. To address this, the existing study adopts a zero-shot transfer strategy: agents are trained on small, tractable instances to learn generalizable exploration heuristics and then deployed to solve larger, unseen instances~\cite{Delgado_Sanchez}.
However, this transfer is often hindered by \textit{anisotropic generalization}~\cite{SEAMS2026}. Since a policy is shaped by its specific training trajectory, it tends to become a ``specialist'' for that configuration while remaining fragile in others. Relying on a single ``best'' policy often fails to address this inherent bias, motivating the need for ensemble methods such as Mixture-of-Experts.

\subsection{Mixture-of-Experts and Gating Strategies}
\label{sec:background_moe}

Mixture-of-Experts (MoE) is an ensemble learning framework designed to decompose complex tasks among specialized models, or \textit{experts} $\{\pi_1, \dots, \pi_m\}$~\cite{jacobs1991adaptive}. A gating network $g(x)$ determines how to combine these experts for a given input $x$. While many modern high-performance architectures, such as Large Language Models, favor a ``hard selection'' (Top-$k$) strategy to activate only a subset of experts for computational efficiency~\cite{shazeer2017outrageously}, this approach can be brittle in decision-making tasks where selection errors lead to irreversible failure.

To provide higher robustness, the ``Soft Mixing'' strategy utilizes the weighted outputs of the entire ensemble simultaneously. Formally, the mixed policy is defined as:
\begin{equation}
    \pi_{\text{soft}}(a|s) = \sum_{i=1}^m g_i(s) \cdot \pi_i(a|s)
\end{equation}
The core advantage of soft mixing lies in its \textit{smoothing effect}. By aggregating the probability distributions of multiple experts, the framework can neutralize the extreme biases or overconfidence of individual models. In the context of safety-critical search and synthesis, this ensemble-based averaging serves as a robust safety net; even if a specialized expert erroneously assigns a low probability to a critical transition due to overfitting or training stochasticity, the collective intelligence of the remaining experts ensures the path remains viable for exploration.

\section{Proposal: Soft Mixture-of-Experts}
\label{sec:proposal}
This section introduces our Soft MoE framework, which mainly consists of the following two phases.

\begin{itemize}
    \item \textbf{Training Phase:} We pre-train a bank of diverse expert policies $\{\pi_1, \dots, \pi_m\}$ on different domain parameter configurations. We then evaluate them offline to construct a \textit{Prior Strength Map}, which quantifies each expert's historical competence across the parameter space using Gaussian-weighted interpolation of solved steps.
    \item \textbf{Synthesis Phase:} When a new synthesis request arrives with parameters $\boldsymbol{\theta}$ and initial state $s_0$, the Gating Network computes a set of mixture weights $g = \{g_1, \dots, g_m\}$. These weights are computed \textit{once} at the start of the episode using both the offline prior and online confidence signals (Entropy and Margin), and are kept fixed to ensure temporal consistency. During exploration, the final policy $\pi_{\text{mix}}$ is synthesized by softly mixing the outputs of all experts according to $g$.
\end{itemize}

\subsection{Training Phase: Prior Strength Construction}
\label{sec:proposal_prior}

The objective of the training phase is to prepare specialized experts and quantify their expected performance across the parameter space.
This phase proceeds in three sequential steps.

First, we train a diverse set of expert policies $\{\pi_1, \dots, \pi_m\}$ using Reinforcement Learning. Each expert is trained under a distinct parameter configuration $\boldsymbol{\theta}_{\text{train}}$ (e.g., specific combinations of traffic volume and altitude levels). This results in a pool of specialists, where each agent is optimized for a specific environmental condition.

Next, we profile each expert by evaluating them offline across a broad range of parameter configurations, including those unseen during training. For every instance where expert $i$ successfully synthesizes a controller, we record the parameter setting $(n_j, k_j)$ and the required expansion steps $L_j$. These records are aggregated into a historical dataset $\mathcal{D}_i = \{(n_j, k_j, L_j)\}_{j}$.

Finally, we construct a map to estimate an expert's competence for any arbitrary query parameter $\boldsymbol{\theta}_q=(n_q, k_q)$ using Gaussian kernel interpolation on $\mathcal{D}_i$.
For a historical sample $j \in \mathcal{D}_i$, the relevance weight $w_j$ is calculated as:
\begin{equation}
    w_j = \exp\left( -0.5 \left( \frac{(n_j - n_q)^2}{\sigma_n^2} + \frac{(k_j - k_q)^2}{\sigma_k^2} \right) \right)
\end{equation}
where $\sigma_n$ and $\sigma_k$ are bandwidth parameters. The estimated step cost $\hat{S}_i(\boldsymbol{\theta}_q)$ is computed as the weighted average of the historical costs $L_j$.
To utilize this as a logic for gating, we standardize the negative estimated cost (since fewer steps imply higher performance):
\begin{equation}
    a_i(\boldsymbol{\theta}) = \frac{-\hat{S}_i(\boldsymbol{\theta}) - \mu}{\sigma + \epsilon}
\end{equation}
This $a_i(\boldsymbol{\theta})$ serves as the \textit{Prior Strength}, representing the baseline confidence in expert $i$ for the target parameters.

\subsection{Synthesis Phase: Prior-Confidence Gating Mechanism}
\label{sec:proposal_gating}
Once the prior knowledge is established, the system handles runtime synthesis requests through three sequential steps: initialization, gating weight computation, and the mixture exploration loop.

Firstly, the phase begins when the system receives a synthesis query consisting of the target domain parameters $\boldsymbol{\theta}_q$ (e.g., specific traffic and altitude constraints) and the initial state of the system $s_0$.

Secondly, before exploration, we compute the mixture weights $g = \{g_1, \dots, g_m\}$. To ensure temporal consistency and reduce overhead, these weights are calculated \textit{once} at the initial state $s_0$ and remain fixed throughout the episode.
The gating logic integrates the offline prior strength $a_i(\boldsymbol{\theta})$ (from Sec.~\ref{sec:proposal_prior}) with online confidence metrics derived from a single forward pass at $s_0$: the expert's Entropy $H_i$ (uncertainty) and Top-2 Margin $M_i$ (decisiveness).
The final gating logit $\ell_i$ is computed as:
\begin{equation}
\label{eq:gating_logit}
    \ell_i = a_i(\boldsymbol{\theta}) - \beta \cdot H_i(s_0) + \gamma \cdot M_i(s_0)
\end{equation}
where $\beta, \gamma \geq 0$ are coefficients. The mixture weights are obtained via softmax: $g_i = \exp(\ell_i) / \sum_{j} \exp(\ell_j)$.

Finally, the system enters the OTF-DCS exploration loop. At each step $t$, instead of selecting a single expert, we employ a \textit{Mixing} strategy. The policy distributions from all experts are aggregated using the fixed weights $g$:
\begin{equation}
\label{eq:mixing}
    \pi_{\text{mix}}(a|s_t) = \sum_{i=1}^m g_i \cdot \pi_i(a|s_t)
\end{equation}
The next transition is selected from this mixed distribution. This mixing provides a smoothing effect, hedging against the risk of overfitting experts while efficiently guiding the search toward the controller synthesis goal.

\begin{algorithm}[t]
\caption{Gating weight computation with Gaussian Prior Interpolation}
\label{alg:moe-gate}
\begin{algorithmic}[1]
\Require target parameters $\boldsymbol{\theta}_q=(n_q, k_q)$
\Require expert history $\mathcal{D}_i = \{(n_j, k_j, L_j)\}_j$ for each expert $i$ \Comment{$L_j$: step cost}
\Require observed initial state $s_0$, coefficients $\beta, \gamma$, bandwidths $\sigma_n, \sigma_k$

\State \textbf{Step 1: Compute One-shot Confidence at $s_0$}
\For{$i = 1,\dots,m$}
  \State $P_i(\cdot|s_0) \gets \text{softmax}(\text{logits}_i(s_0)/T)$
  \State $H_i \gets -\sum_a P_i(a)\log P_i(a)$ \Comment{Entropy (Uncertainty)}
  \State $M_i \gets P_i^{(1)} - P_i^{(2)}$ \Comment{Top-2 Margin (Decisiveness)}
\EndFor

\State \textbf{Step 2: Compute Gaussian-Weighted Prior Strength}
\For{$i = 1,\dots,m$}
  \State Initialize weighted sum $W_{\text{sum}} \gets 0$, cost sum $C_{\text{sum}} \gets 0$
  \For{each sample $j \in \mathcal{D}_i$ (solved instances)}
    \State $d^2 \gets ((n_j - n_q)/\sigma_n)^2 + ((k_j - k_q)/\sigma_k)^2$
    \State $w_j \gets \exp(-0.5 \cdot d^2)$ \Comment{Gaussian kernel weight}
    \State $W_{\text{sum}} \gets W_{\text{sum}} + w_j, \quad C_{\text{sum}} \gets C_{\text{sum}} + w_j \cdot L_j$
  \EndFor
  \If{$W_{\text{sum}} > \epsilon$}
    \State $\hat{S}_i \gets C_{\text{sum}} / W_{\text{sum}}$ \Comment{Estimated step cost at $\boldsymbol{\theta}_q$}
  \Else
    \State $\hat{S}_i \gets \infty$ \Comment{No nearby history}
  \EndIf
\EndFor
\State $a_i \gets \text{Standardize}(-\hat{S}_i)$ \Comment{Normalize negative cost (fewer is better)}

\State \textbf{Step 3: Compute Mixture Weights}
\State $\ell_i \gets a_i - \beta H_i + \gamma M_i$ \Comment{Combine Prior and Confidence}
\State $g_i \gets \exp(\ell_i) / \sum_j \exp(\ell_j)$
\State \Return $g = (g_1, \dots, g_m)$
\end{algorithmic}
\end{algorithm}

\begin{algorithm}[t]
\caption{MoE-guided Exploration Episode (Soft vs. Hard)}
\label{alg:moe-episode}
\begin{algorithmic}[1]
\Require parameters $\boldsymbol{\theta}$, experts $\{\pi_i\}$, budget $B$, Mode $\in \{\text{Soft, Hard}\}$
\Ensure cumulative return $R$, solved flag $\mathit{done}$

\State Initialize env, observe $s_0$
\State $g \gets \textsc{Gating}(\boldsymbol{\theta}, \{\pi_i\}, s_0)$ \Comment{Compute weights once using Alg.~\ref{alg:moe-gate}}

\If{Mode is \textbf{Hard}}
  \State $k^* \gets \arg\max_i g_i$ \Comment{Select single best expert}
  \State $\pi_{\text{active}} \gets \pi_{k^*}$ \Comment{Fix expert for the episode}
\EndIf

\State $t \gets 0$
\While{$t < B$ and not $\mathit{done}$}
  \If{Mode is \textbf{Hard}}
    \State $P_{\text{final}}(\cdot) \gets \pi_{\text{active}}(\cdot|s_t)$ \Comment{Use selected expert only}
  \ElsIf{Mode is \textbf{Soft}}
    \For{$i = 1,\dots,m$}
      \State $P_i(\cdot) \gets \pi_i(\cdot|s_t)$
    \EndFor
    \State $P_{\text{final}}(\cdot) \gets \sum_{i=1}^m g_i \cdot P_i(\cdot)$ \Comment{Weighted Mixture of all experts}
  \EndIf

  \State $a_t \gets \arg\max_a P_{\text{final}}(a)$ \Comment{Greedy selection from mixed policy}
  \State $(s_{t+1}, r_t, \mathit{done}) \gets \text{Env.step}(a_t)$
  \State $t \gets t+1$
\EndWhile
\end{algorithmic}
\end{algorithm}

\section{Evaluation}
\label{sec:evaluation}

This section evaluates the proposed Soft-MoE framework. We address the following research questions (RQs) to verify the effectiveness, theoretical validity, and efficiency of our approach:

\begin{itemize}
    \item \textbf{RQ1: Anisotropic Generalization.} To what extent does the training environment configuration limit the generalization capability of single RL exploration policies?

    \item \textbf{RQ2: Effectiveness.} How effectively does the proposed MoE framework leverage the complementary strengths of diverse experts to expand the solvable parameter space?

    \item \textbf{RQ3: Computation Cost.} What is the computational overhead introduced by the MoE framework, and is this cost justifiable given the improvements in robustness?
\end{itemize}

\subsection{Experiment Setup}
\label{sec:eval_settings}

\textit{Benchmark}. We employ the Air Traffic (AT) management system benchmark~\cite{CIOLEK2020}, a standard problem in DCS. The complexity of this domain is characterized by two parameters: $n$, the number of airplanes requesting to land, and $k$, the number of available altitude levels. The goal is to synthesize a controller that guarantees collision avoidance.

\textit{Training Configuration}.
To prepare multiple experts, we trained a diverse set of expert policies. Instead of training on a single configuration, we selected six distinct parameter settings for training: $\theta(n,k) \in \{2,3\} \times \{1,2,3\}$ (excluding the trivial case $n=1$). For each setting, we trained 10 independent models with different random seeds, resulting in a total pool of 60 expert policies. Each expert was trained for 1,000 episodes with a computational budget of 10,000 transitions per episode.

\textit{Evaluation Protocol}.
The evaluation is conducted in a zero-shot transfer manner. We test the trained experts and the proposed MoE framework on a broad range of unseen parameter configurations, specifically $1 \le n, k \le 15$, creating a grid of 225 test instances.
An instance is considered ``solved'' if a valid non-blocking controller is synthesized within a hard budget of 10,000 expanded transitions.
For the MoE framework, we evaluate mixtures of increasing sizes, denoted as T1 to T6, where T$x$ combines $x$ experts. The hyperparameters for the gating mechanism are set to $\beta = \gamma = 1.0$, and the temperature $T = 2.0$.

\subsection{RQ1: Anisotropic Generalization}
\label{sec:eval_rq1}

\begin{figure}[h!tb]
  \centering
  \includegraphics[width=\linewidth]{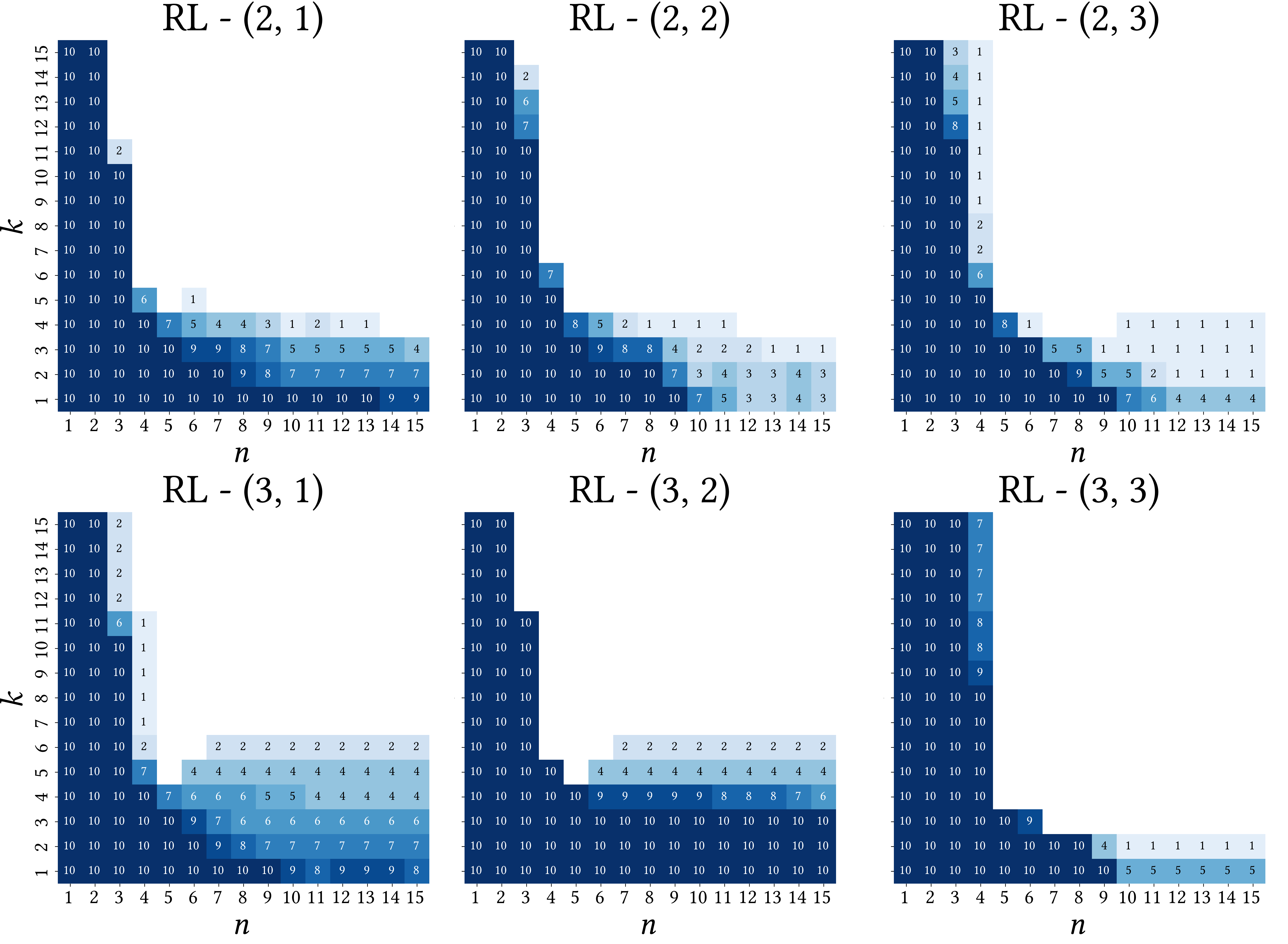}
  \caption{Zero-Shot Generalization of Single RL Experts in Success Rates.}
  \label{fig:heatmap_rl_instance}
\end{figure}

\begin{figure}[h!tb]
  \centering
  \includegraphics[width=\linewidth]{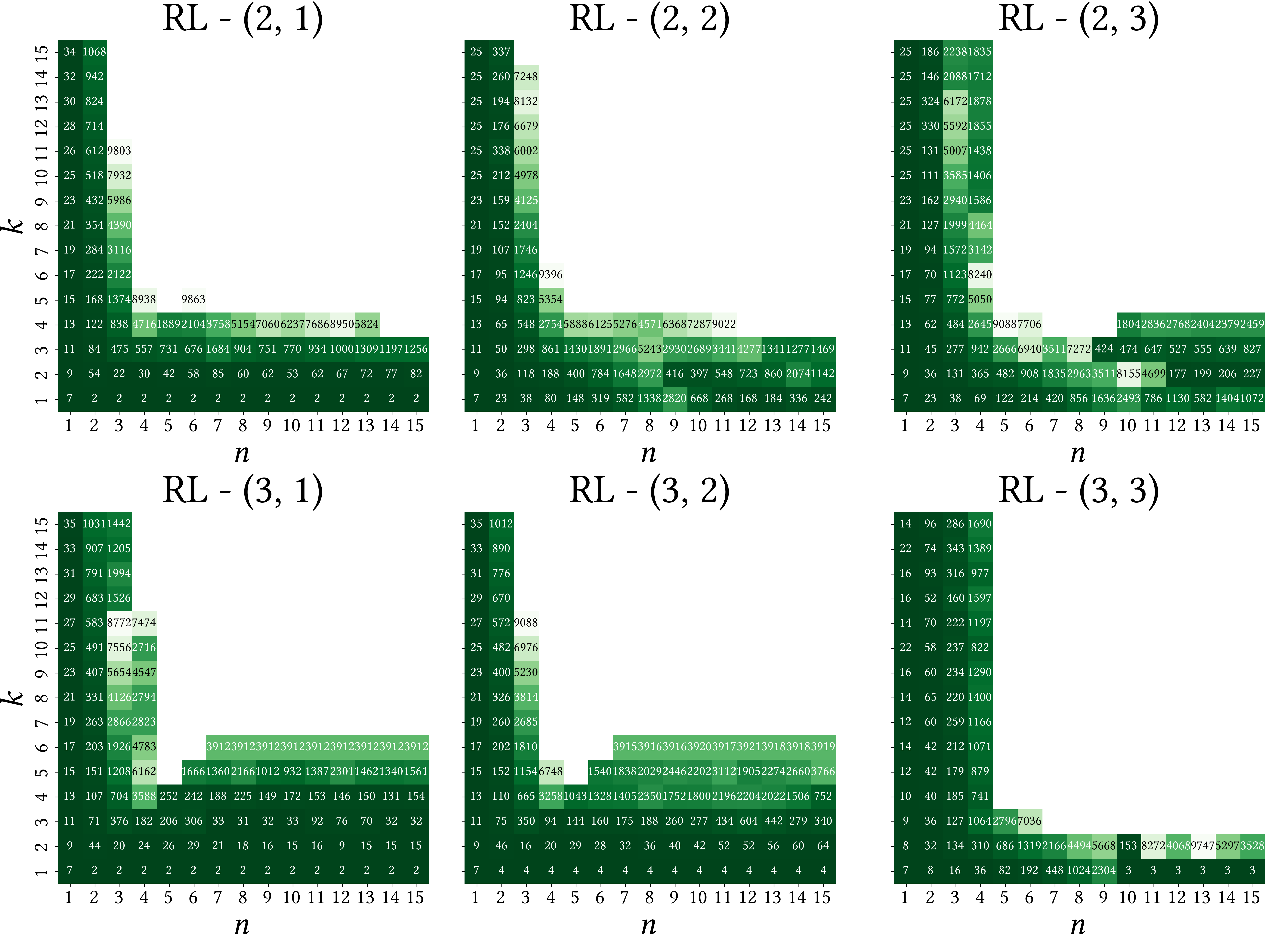}
  \caption{Zero-Shot Generalization of Single RL Experts in Exploration Efficiency (Number of Steps).}
  \label{fig:heatmap_rl_step}
\end{figure}

To understand the baseline performance and the necessity of an ensemble approach, we first analyze the generalization capability of single RL experts.
Fig.~\ref{fig:heatmap_rl_instance} and Fig.~\ref{fig:heatmap_rl_step} illustrate the success rates and number of expansion steps required of six representative experts across the $15 \times 15$ parameter grid.
We observe a distinct anisotropic generalization pattern: each expert exhibits a high success rate (dark blue regions) only within a narrow corridor or "diagonal" surrounding its training parameters $(n, k)$.
As the test instances deviate from the training distribution---particularly when the system scale $n$ or the constraints $k$ increase---the performance drops sharply, often resulting in complete synthesis failure (light-colored regions).

These results shows that single RL policies suffer from inherent bias derived from their unique training trajectories.
The sparse and non-overlapping nature of the high-performance regions across different experts (as seen in the comparison between RL-(2, 1) and RL-(3, 3)) suggests that different experts possess complementary expertise.
In summary, this observation provides a strong motivation for an MoE-based approach.

\subsection{RQ2: Effectiveness}
\label{sec:eval_rq2}

\begin{figure}[h!tb]
  \centering
  \includegraphics[width=\linewidth]{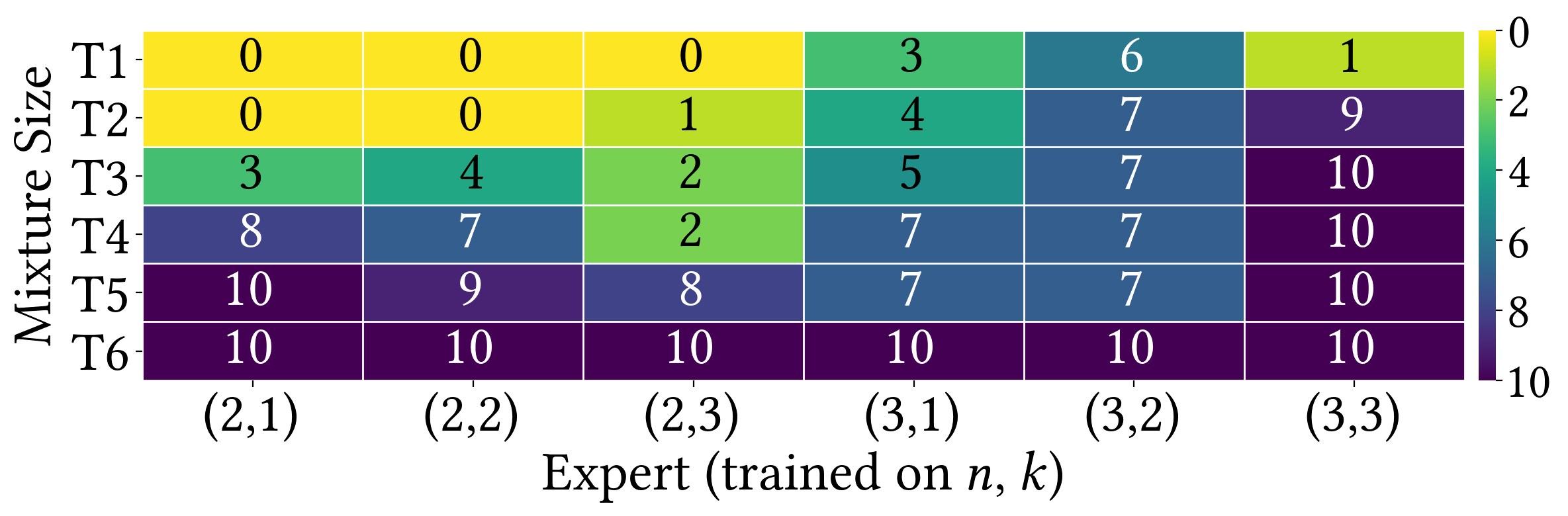}
  \caption{Dominant Expert Selection Map. }
  \label{fig:selected_teachers}
\end{figure}

Figure~\ref{fig:selected_teachers} illustrates the \emph{selection order} of experts as the mixture size grows. We observe a clear two-stage tendency: in the first round, experts trained at $(3,1)$ and $(3,2)$ are most frequently selected as the initial backbone. In the second round, the expert $(3,3)$---which is complementary to them---is added most often. This selection pattern closely matches the anisotropic generalization trends observed earlier, where $(3,1)/(3,2)$ provide broader baseline coverage while $(3,3)$ compensates for their fragile regions.

\begin{figure}[h!tb]
  \centering
  \includegraphics[width=\linewidth]{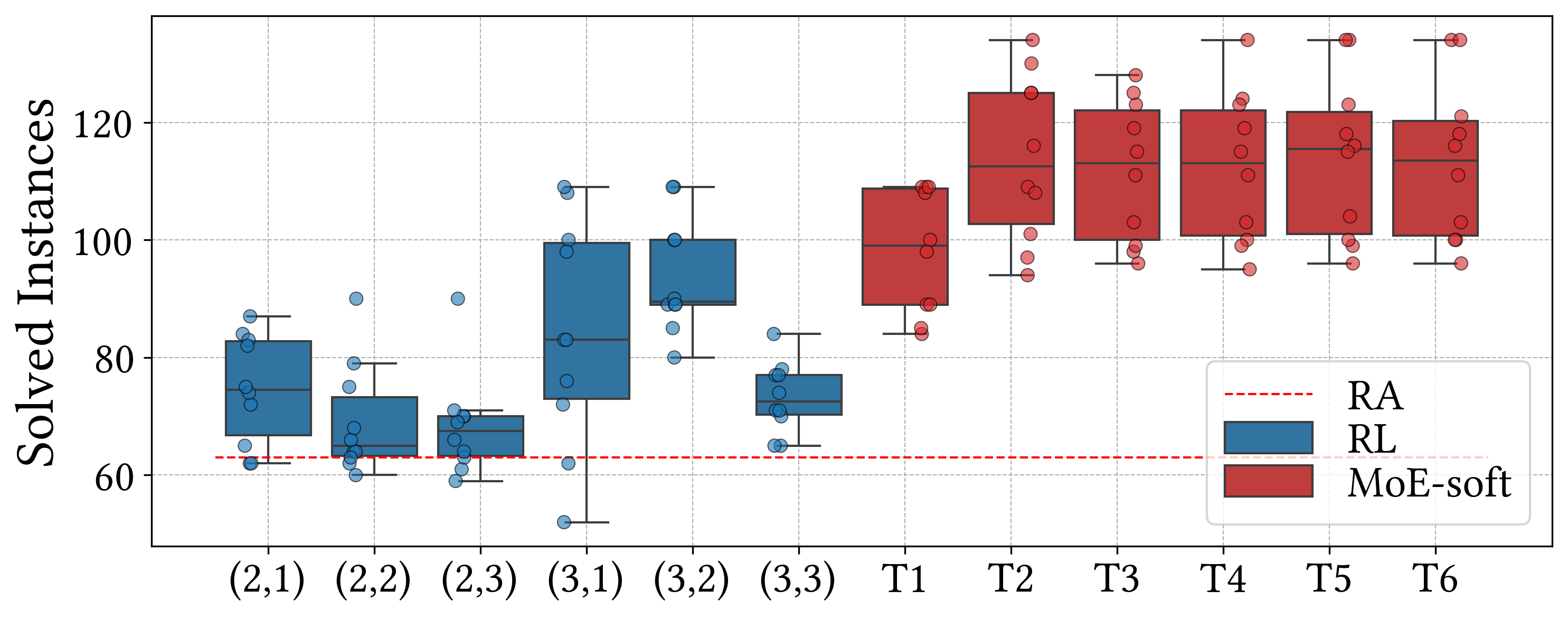}
  \caption{Overall Performance Comparison.}
  \label{fig:boxplot_stability}
\end{figure}

\begin{figure}[h!tb]
  \centering
  \includegraphics[width=\linewidth]{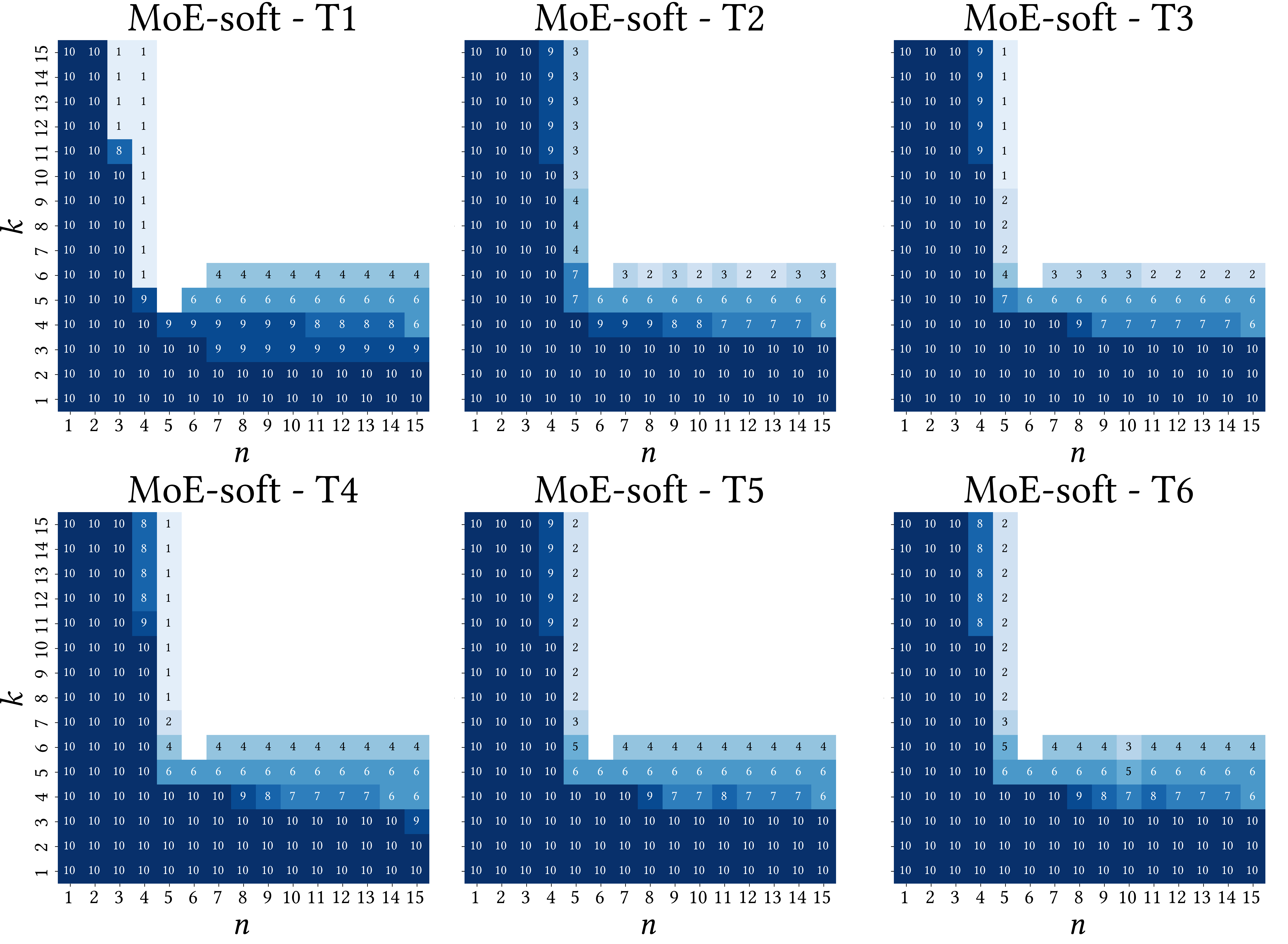}
  \caption{Success Heatmaps of MoE Mixtures (T1--T6).}
  \label{fig:heatmap_moe_instance}
\end{figure}

\begin{figure}[h!tb]
  \centering
  \includegraphics[width=\linewidth]{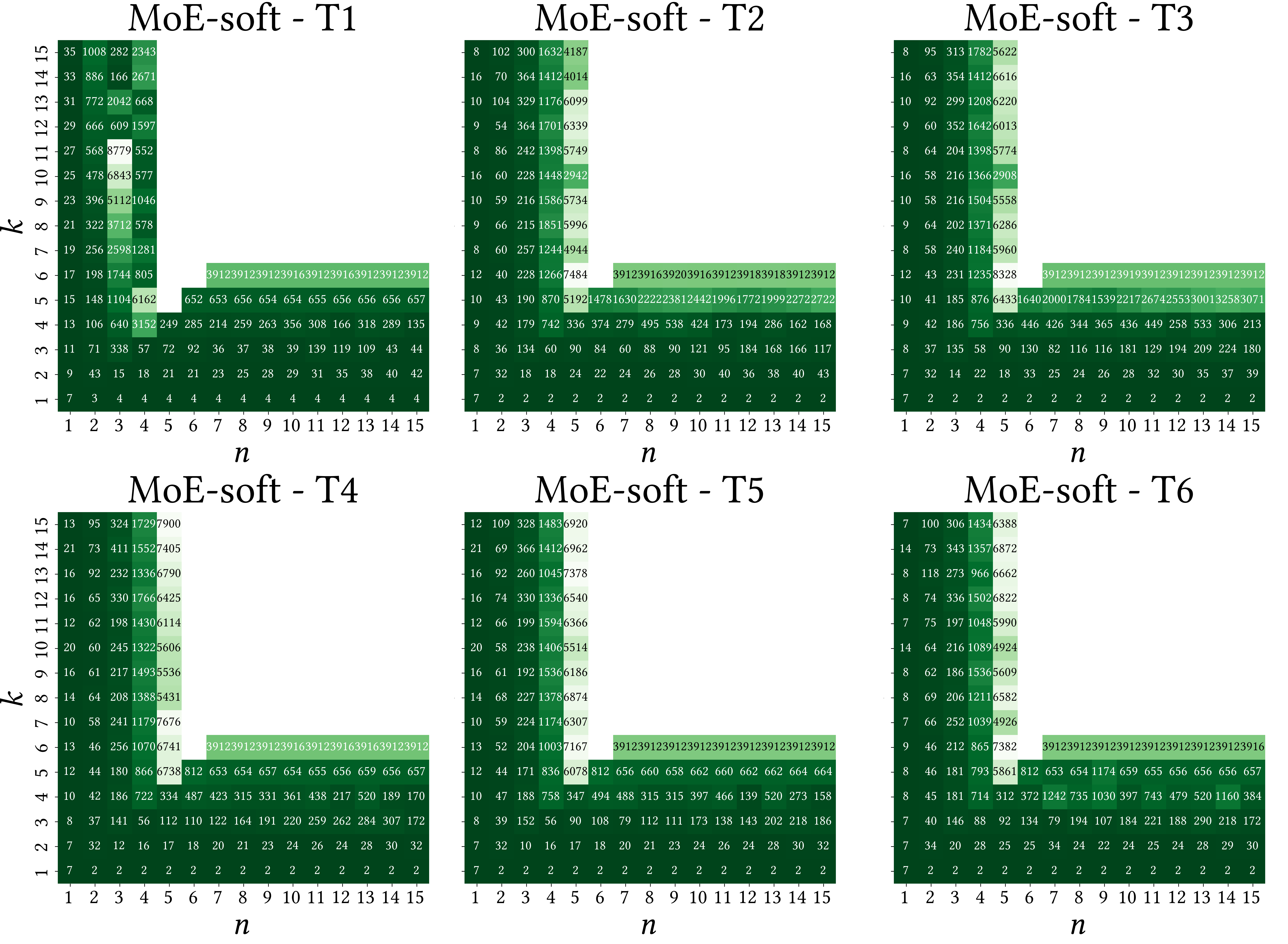}
  \caption{Exploration Efficiency of MoE Mixtures (T1--T6).}
  \label{fig:heatmap_moe_step}
\end{figure}

The success heatmaps in Fig.~\ref{fig:heatmap_moe_instance} and the step-cost heatmaps in Fig.~\ref{fig:heatmap_moe_step} show, respectively, how the solvable parameter space expands and how exploration efficiency changes across mixtures (T1--T6). Consistent with the overall performance trend in Fig.~\ref{fig:boxplot_stability}, the improvement is most pronounced from T1 to T2, while the gains beyond that become marginal.

Figure~\ref{fig:boxplot_stability} compares the number of solved instances across single experts and MoE mixtures. Single RL experts exhibit substantial variance and limited coverage, reflecting anisotropic generalization. In contrast, MoE configurations (T1--T6) consistently outperform individual experts. The performance gain is most significant from T1 to T3, after which improvements gradually saturate.

\subsection{RQ3: Computation Cost}
\label{sec:eval_rq3}

\begin{figure}[h!tb]
  \centering
  \includegraphics[width=\linewidth]{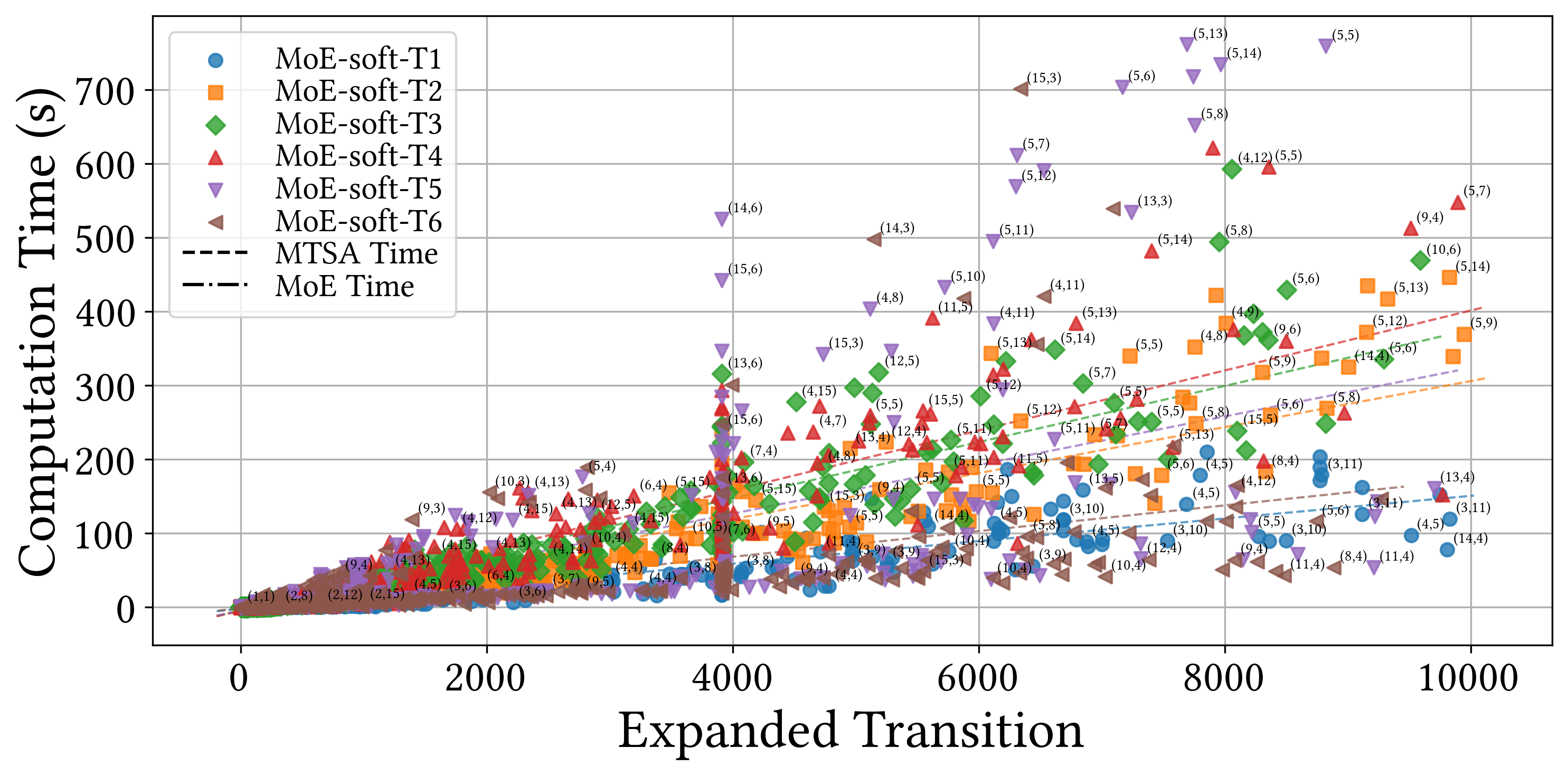}
  \caption{Computational Overhead Scaling.}
  \label{fig:time_scaling}
\end{figure}

Finally, we analyze the computational overhead introduced by the ensemble approach.
Figure~\ref{fig:time_scaling} analyzes the computational overhead. As expected, the per-step inference cost increases with the number of experts because multiple neural networks must be evaluated.
However, interestingly, the total synthesis time does not scale linearly with the ensemble size (i.e., T6 is not simply six times slower than T1). This is because larger mixtures provide superior heuristics, which significantly reduce the total number of expanded transitions required to find a solution (as observed in Fig.~\ref{fig:heatmap_moe_step}).
In other words, the increased inference cost per step is partially offset by the reduction in the total number of steps.

\subsection{Discussion and Limitations}

Our results suggest that the effectiveness of Soft-MoE primarily comes from \emph{structured complementarity} rather than merely increasing ensemble size. The selection-order analysis (Fig.~\ref{fig:selected_teachers}) indicates that the gating tends to first anchor on broadly generalizing experts (e.g., $(3,1)$ and $(3,2)$), and then incorporate the most complementary expert (often $(3,3)$), which matches the anisotropic patterns observed in RQ1. This explains why the most visible gain appears early (T1$\rightarrow$T2): once the key complementary directions in the parameter space are covered, additional experts mainly provide redundancy, leading to diminishing returns in both success coverage (Fig.~\ref{fig:heatmap_moe_instance}) and efficiency (Fig.~\ref{fig:heatmap_moe_step}). From a practical perspective, this implies that a small mixture may already capture most of the robustness benefit, while larger mixtures are primarily useful when maximizing coverage is prioritized over runtime overhead.

Regarding limitations, first, our gating weights are computed once at $s_0$ and then fixed throughout the episode (Alg.~\ref{alg:moe-gate}), which improves temporal consistency and reduces overhead, but may be suboptimal when the search dynamics change substantially as exploration progresses. A more context-aware gating mechanism that adapts online could further improve robustness, yet it also risks oscillatory behavior and extra computation.

Second, MoE incurs non-negligible inference overhead because it evaluates multiple experts per step (Sec.~\ref{sec:eval_rq3}). Although the total runtime does not scale linearly due to fewer expanded transitions, the approach can still be too slow for highly volatile environments where re-synthesis must complete within a strict reaction time. In such settings, Soft-MoE is better viewed as a component within a broader adaptation pipeline (e.g., combining with a safe/degraded fallback controller) while the new controller is being synthesized.

Third, our evaluation is centered on a two-dimensional parameter space $(n,k)$ and a single benchmark domain (AT). This simplifies both expert coverage and prior construction, but real-world settings can involve higher-dimensional or partially observed parameters, where expert training and prior mapping face the curse of dimensionality. Future work should investigate (i) principled expert sampling and coverage planning in high-dimensional spaces, (ii) more compact representations of prior strength (e.g., learned surrogates), and (iii) distillation/compression to reduce inference cost while preserving the ensemble's robustness.

\subsection{Threats to Validity}

\textit{Internal validity.} Our results may be influenced by sensitivity to hyperparameters (e.g., $\beta,\gamma,T,\sigma_n,\sigma_k$) and training stochasticity across seeds. In addition, the fixed transition budget and the "solved" criterion can affect the observed saturation point, and the specific expert pool trained on ${2,3}\times{1,2,3}$ may bias the selection order.

\textit{External validity.} Our evaluation is limited to the Air Traffic benchmark and a two-dimensional parameter space $(n,k)$. Other DCS domains (with different interaction/topology or contention patterns) may exhibit different anisotropic behaviors and thus different MoE benefits. Moreover, the prior-strength profiling assumption may be weaker in higher-dimensional, continuous, or non-stationary settings, where prior coverage is sparse. Broader validation across domains and parameter regimes is required to generalize these findings.

\section{Related Work}
\label{sec:related}

\subsection{Computation Efficiency for Controller Synthesis}
Given the state-explosion nature of controller synthesis, the reduction of the computational cost of controller synthesis has been widely studied. Existing research can generally be categorized into the following approaches.
The first category is Directed Controller Synthesis (DCS) \cite{Ciolek:2016:DCS, Delgado_Sanchez, ubukata2025graph}, which reduces the computational space required by incrementally exploring and constructing the non-violating and non-blocking state space starting from the initial state. Once a control strategy that satisfies the specifications is found, the exploration terminates, thereby avoiding an exhaustive search over the entire game space.
The second category is known as Compositional Controller Synthesis, whose specific implementation varies depending on whether it targets a safety property \cite{stepwise,ishimizu2024towards} or a GR(1) property \cite{10.1007/978-3-031-98685-7_10}.
Besides, there are also a few other niche approaches. For instance, \cite{jialong_RE22} reduces computation time by model abstraction, and \cite{jialong_EUC20} adopts incremental analysis for re-synthesis in a changed environment.
Among the above approaches, this paper falls under the first category and specifically addresses the issues observed in the existing RL-based methods, to further improve the efficiency of DCS.

\subsection{Mixture-of-Experts for Reinforcement Learning}
Recently, MoE has gained attention in RL \cite{hendawy2024multitask,VASIC202234,DBLP:conf/iclr/ChowTNGRGB23,wu2025mixtureofexpertsmeetsincontextreinforcement,wang2022learning}, driven by two main motivations.
First, RL often suffers from non-stationarity caused by task switching or multi-task RL (MTRL), i.e., a single model typically struggles to handle multiple tasks simultaneously, whereas the modular structure of MoE offers clear advantages. For instance, \cite{willi2024mixtureexpertsmixturerl} applies MoE to jointly train multiple Atari games and sequential tasks. \cite{huang2025mentormixtureofexpertsnetworktaskoriented} utilizes MoE architecture to enable different experts to learn distinct types of action policies for robotic skills.
Second, MoE can improve the scalability of RL. For instance, \cite{10.5555/3692070.3693633} embeds MoE modules into value function networks and demonstrates significant performance gains in deep RL algorithms as the model size increases.
Unlike the two approaches discussed above, this paper employs MoE, intending to address the issue of anisotropic generalization arising from differences in learning environments and training histories. This challenge is mainly situated within the various environments within one single task domain, rather than across multiple tasks. To tackle this, our MoE design not only considers the input features (which are the primary input in the above studies) but also incorporates each expert's prior strength and confidence.

\section{Conclusion}
\label{sec:conclusion}

This paper investigated the limitations of RL-based exploration policies in on-the-fly directed controller synthesis, namely, anisotropic generalization across domain parameters. Through a systematic evaluation, we showed that single experts tend to generalize well only in narrow regions of the parameter space, and their high-performance regions are sparse and complementary. Motivated by this observation, we proposed Soft-MoE, a robust exploration framework that integrates multiple RL experts via a prior-confidence gating mechanism and soft mixing. The gating combines design-time prior strength (constructed by Gaussian interpolation over historical solved steps) with runtime one-shot confidence signals (entropy and margin) at the initial state. Experimental results on the Air Traffic benchmark demonstrate that Soft-MoE substantially expands the solvable space compared to any single expert.

In future work, we plan to extend our research in the following two directions.
First, we will develop a more context-aware adaptive gating mechanism that can dynamically adjust expert contributions during the exploration process, rather than fixing them at the initial state, to better respond to shifting state-space characteristics as the search progresses.
Second, we plan to explore the use of model distillation to compress the collective intelligence of the MoE ensemble into a single, efficient student policy, thereby significantly reducing the computational overhead and latency required for real-time controller synthesis in resource-constrained environments.

\bibliography{ref}

@inproceedings{Delgado_Sanchez,
  author    = {Tom{\'{a}}s Delgado and Marco S{\'{a}}nchez Sorondo and V{\'{i}}ctor Braberman and Sebasti{\'{a}}n Uchitel},
  title     = {Exploration Policies for On-the-Fly Controller Synthesis: {A} Reinforcement Learning Approach},
  booktitle = {Proceedings of the 33rd International Conference on Automated Planning and Scheduling (ICAPS)},
  pages     = {569--577},
  year      = {2023},
  publisher = {AAAI Press},
}

@ARTICLE{CIOLEK2020,
  author={Ciolek, Daniel Alfredo and Braberman, Victor and D’Ippolito, Nicolás and Sardiña, Sebastián and Uchitel, Sebastián},
  journal={IEEE Transactions on Automatic Control}, 
  title={Compositional Supervisory Control via Reactive Synthesis and Automated Planning}, 
  year={2020},
  volume={65},
  number={8},
  pages={3502-3516},
  doi={10.1109/TAC.2019.2948270}
}

@ARTICLE{Wonham,
  author={Ramadge, P.J.G. and Wonham, W.M.},
  journal={Proceedings of the IEEE}, 
  title={The control of discrete event systems}, 
  year={1989},
  volume={77},
  number={1},
  pages={81-98},
  doi={10.1109/5.21072}}

@article{Tei,
title = "Dynamic Update of Discrete Event Controllers",
author = "Leandro Nahabedian and Victor Braberman and Nicolas Dippolito and Shinichi Honiden and Jeff Kramer and Kenji Tei and Sebastian Uchitel",
note = "Publisher Copyright: {\textcopyright} 1976-2012 IEEE.",
year = "2020",
month = nov,
day = "1",
doi = "10.1109/TSE.2018.2876843",
language = "English",
volume = "46",
pages = "1220--1240",
journal = "IEEE Transactions on Software Engineering",
issn = "0098-5589",
publisher = "Institute of Electrical and Electronics Engineers Inc.",
number = "11",
}

@misc{benchmark,
  title = {{Benchmark}},
  howpublished = 
  "\url{https://bitbucket.org/lnahabedian/mtsa/src/master/maven-root/mtsa/src/test/benchmarks/OTF-NonBlockingBenchmark/fsp/}",
  year = {2021}, 
}

@InProceedings{10.1007/3-540-48119-2_15,
author="Tripakis, Stavros
and Altisen, Karine",
editor="Wing, Jeannette M.
and Woodcock, Jim
and Davies, Jim",
title="On-the-Fly Controller Synthesis for Discrete and Dense-Time Systems",
booktitle="FM'99 --- Formal Methods",
year="1999",
publisher="Springer Berlin Heidelberg",
address="Berlin, Heidelberg",
pages="233--252",
isbn="978-3-540-48119-5",
  doi={doi.org/10.1007/3-540-48119-2_15}}

@book{cassandras2008introduction,
  title={Introduction to Discrete Event Systems},
  author={Cassandras, Christos G. and Lafortune, St{\'e}phane},
  year={2008},
  publisher={Springer},
  address={New York, NY},
  edition={2nd},
  doi={10.1007/978-0-387-68612-7},
  isbn={978-0-387-68612-7}
}

@InProceedings{10.1007/978-3-031-98685-7_10,
author="Gagliardi, Hernan
and Braberman, Victor
and Uchitel, Sebastian",
editor="Piskac, Ruzica
and Rakamari{\'{c}}, Zvonimir",
title="Scaling GR(1) Synthesis via a Compositional Framework for LTL Discrete Event Control",
booktitle="Computer Aided Verification",
year="2025",
publisher="Springer Nature Switzerland",
address="Cham",
pages="201--223",
isbn="978-3-031-98685-7"
}

@book{Cassandras:2006:DES,
	title = {Introduction to Discrete Event Systems},
	author = {C. G. Cassandras and S. Lafortune},
	publisher = {Springer-Verlag},
	year = {2006}
}

@inproceedings{Ciolek:2016:DCS,
  title = {{Directed Controller Synthesis of discrete event systems: Taming composition with heuristics}},
  author = {D. Ciolek and V. Braberman and N. D'Ippolito and S. Uchitel},
  booktitle = {Proc. of the IEEE Conf. on Decision and Control},
  series = {CDC},
  pages = {4764--4769},
  year = {2016}
}

@article{CiolekDCS,
title = {On-the-fly informed search of non-blocking directed controllers},
journal = {Automatica},
volume = {147},
pages = {110731},
year = {2023},
issn = {0005-1098},
author = {Daniel Ciolek and Matias Duran and Florencia Zanollo and Nicolas Pazos and Julián Braier and Victor Braberman and Nicolas D’Ippolito and Sebastian Uchitel}
}

@inproceedings{jialong_RE22,
	title        = {Done is better than perfect: Iterative Adaptation via Multi-grained Requirement Relaxation},
	author       = {Li, Jialong and Tei, Kenji},
	year         = 2022,
	booktitle    = {2022 IEEE 30th International Requirements Engineering Conference (RE)},
	volume       = {},
	number       = {},
	pages        = {288-294},
	doi          = {10.1109/RE54965.2022.00043}
}

@inproceedings{jialong_EUC20,
	title        = {Efficient Difference Analysis Algorithm for Runtime Requirement Degradation under System Functional Fault},
	author       = {Li, Jialong and Aizawa, Kazuya and Tei, Kenji and Honiden, Shinichi},
	year         = 2020,
	booktitle    = {2020 IEEE 18th International Conference on Embedded and Ubiquitous Computing (EUC)},
	volume       = {},
	number       = {},
	pages        = {33-40}
}

@article{stepwise,
  author       = {Takuto Yamauchi and Kenji Tei},
  title        = {An Analysis Space Reduction for Discrete Controller Synthesis by Stepwise Partial Synthesis},
  journal      = {IEICE Transactions D},
  volume       = {J106-D},
  number       = {4},
  pages        = {218-230},
  year         = {2023},
  month        = {April},
  issn         = {1881-0225},
  doi          = {10.14923/transinfj.2022PDP0022}
}

@inproceedings{10.1145/3643915.3644095,
author = {Carwehl, Marc and Imrie, Calum and Vogel, Thomas and Rodrigues, Gena\'{\i}na and Calinescu, Radu and Grunske, Lars},
title = {Formal Synthesis of Uncertainty Reduction Controllers},
year = {2024},
isbn = {9798400705854},
doi = {10.1145/3643915.3644095},
booktitle = {Proceedings of the 19th International Symposium on Software Engineering for Adaptive and Self-Managing Systems (SEAMS)},
pages = {2–13},
numpages = {12}
}

@INPROCEEDINGS{10174043,
  author={Buckworth, Titus and Alrajeh, Dalal and Kramer, Jeff and Uchitel, Sebastian},
  booktitle={2023 IEEE/ACM 18th Symposium on Software Engineering for Adaptive and Self-Managing Systems (SEAMS)}, 
  title={Adapting Specifications for Reactive Controllers}, 
  year={2023},
  volume={},
  number={},
  pages={1-12},
  doi={10.1109/SEAMS59076.2023.00012}}

@inproceedings{10.1145/2897053.2897056,
author = {Nahabedian, L. and Braberman, V. and D'Ippolito, N. and Honiden, S. and Kramer, J. and Tei, K. and Uchitel, S.},
title = {Assured and correct dynamic update of controllers},
year = {2016},
isbn = {9781450341875},
doi = {10.1145/2897053.2897056},
booktitle = {Proceedings of the 11th International Symposium on Software Engineering for Adaptive and Self-Managing Systems (SEAMS)},
pages = {96–107},
numpages = {12},
}

@inproceedings{ishimizu2024towards,
  author={Ishimizu, Yusei and Li, Jialong and Yamauchi, Takuto and Chen, Sinan and Cai, Jinyu and Hirano, Takanori and Tei, Kenji},
  booktitle={2024 IEEE International Conference on Consumer Electronics-Asia (ICCE-Asia)}, 
  title={Towards Efficient Discrete Controller Synthesis: Semantics-Aware Stepwise Policy Design via LLM}, 
  year={2024},
  volume={},
  number={},
  pages={1-4},
  doi={10.1109/ICCE-Asia63397.2024.10773792}}

@inproceedings{10.5555/3692070.3693633,
author = {Obando-Ceron, Johan and Sokar, Ghada and Willi, Timon and Lyle, Clare and Farebrother, Jesse and Foerster, Jakob and Dziugaite, Karolina and Precup, Doina and Castro, Pablo Samuel},
title = {Mixtures of experts unlock parameter scaling for deep RL},
year = {2024},
\abstract = {The recent rapid progress in (self) supervised learning models is in large part predicted by empirical scaling laws: a model's performance scales proportionally to its size. Analogous scaling laws remain elusive for reinforcement learning domains, however, where increasing the parameter count of a model often hurts its final performance. In this paper, we demonstrate that incorporating Mixture-of-Expert (MoE) modules, and in particular Soft MoEs (Puigcerver et al., 2023), into value-based networks results in more parameter-scalable models, evidenced by substantial performance increases across a variety of training regimes and model sizes. This work thus provides strong empirical evidence towards developing scaling laws for reinforcement learning. We make our code publicly available.},
booktitle = {Proceedings of the 41st International Conference on Machine Learning (ICML)},
articleno = {1563},
numpages = {21},
location = {Vienna, Austria},
}

@misc{willi2024mixtureexpertsmixturerl,
      title={Mixture of Experts in a Mixture of RL settings}, 
      author={Timon Willi and Johan Obando-Ceron and Jakob Foerster and Karolina Dziugaite and Pablo Samuel Castro},
      year={2024},
      eprint={2406.18420},
      archivePrefix={arXiv},
      primaryClass={cs.LG},
      url={https://arxiv.org/abs/2406.18420}, 
}

@misc{huang2025mentormixtureofexpertsnetworktaskoriented,
      title={MENTOR: Mixture-of-Experts Network with Task-Oriented Perturbation for Visual Reinforcement Learning}, 
      author={Suning Huang and Zheyu Zhang and Tianhai Liang and Yihan Xu and Zhehao Kou and Chenhao Lu and Guowei Xu and Zhengrong Xue and Huazhe Xu},
      year={2025},
      eprint={2410.14972},
      archivePrefix={arXiv},
      primaryClass={cs.RO},
      url={https://arxiv.org/abs/2410.14972}, 
}

@ARTICLE{jacobs1991adaptive,
  author={Jacobs, Robert A. and Jordan, Michael I. and Nowlan, Steven J. and Hinton, Geoffrey E.},
  journal={Neural Computation}, 
  title={Adaptive Mixtures of Local Experts}, 
  year={1991},
  volume={3},
  number={1},
  pages={79-87},
  doi={10.1162/neco.1991.3.1.79}}

@misc{shazeer2017outrageously,
      title={Outrageously Large Neural Networks: The Sparsely-Gated Mixture-of-Experts Layer}, 
      author={Noam Shazeer and Azalia Mirhoseini and Krzysztof Maziarz and Andy Davis and Quoc Le and Geoffrey Hinton and Jeff Dean},
      year={2017},
      eprint={1701.06538},
      archivePrefix={arXiv},
      primaryClass={cs.LG},
      url={https://arxiv.org/abs/1701.06538}, 
}

@ARTICLE{8876664,
  author={Ciolek, Daniel Alfredo and Braberman, Victor and D’Ippolito, Nicolás and Sardiña, Sebastián and Uchitel, Sebastián},
  journal={IEEE Transactions on Automatic Control}, 
  title={Compositional Supervisory Control via Reactive Synthesis and Automated Planning}, 
  year={2020},
  volume={65},
  number={8},
  pages={3502-3516},
  doi={10.1109/TAC.2019.2948270}}

@inproceedings{10.1145/2897053.2903381,
author = {Uchitel, Sebastian and Braberman, Victor A. and D'Ippolito, Nicolas},
title = {Runtime controller synthesis for self-adaptation: be discrete!},
year = {2016},
isbn = {9781450341875},
doi = {10.1145/2897053.2903381},
booktitle = {Proceedings of the 11th International Symposium on Software Engineering for Adaptive and Self-Managing Systems (SEAMS)},
pages = {1–3},
numpages = {3},
location = {Austin, Texas},
}

@inproceedings{
hendawy2024multitask,
title={Multi-Task Reinforcement Learning with Mixture of Orthogonal Experts},
author={Ahmed Hendawy and Jan Peters and Carlo D'Eramo},
booktitle={The Twelfth International Conference on Learning Representations (ICLR)},
year={2024}}

@article{VASIC202234,
title = {MoËT: Mixture of Expert Trees and its application to verifiable reinforcement learning},
journal = {Neural Networks},
volume = {151},
pages = {34-47},
year = {2022},
issn = {0893-6080},
doi = {https://doi.org/10.1016/j.neunet.2022.03.022},
author = {Marko Vasić and Andrija Petrović and Kaiyuan Wang and Mladen Nikolić and Rishabh Singh and Sarfraz Khurshid},
}

@inproceedings{DBLP:conf/iclr/ChowTNGRGB23,
  author={Yinlam Chow and Aza Tulepbergenov and Ofir Nachum and Dhawal Gupta and Moonkyung Ryu and Mohammad Ghavamzadeh and Craig Boutilier},
  title={A Mixture-of-Expert Approach to RL-based Dialogue Management},
  year={2023},
  booktitle={The Eleventh International Conference on Learning Representations (ICLR)},
}

@misc{wu2025mixtureofexpertsmeetsincontextreinforcement,
      title={Mixture-of-Experts Meets In-Context Reinforcement Learning}, 
      author={Wenhao Wu and Fuhong Liu and Haoru Li and Zican Hu and Daoyi Dong and Chunlin Chen and Zhi Wang},
      year={2025},
      eprint={2506.05426},
      archivePrefix={arXiv},
      primaryClass={cs.LG},
      url={https://arxiv.org/abs/2506.05426}, 
}

@inproceedings{
    wang2022learning,
    title={Learning Expressive Meta-Representations with Mixture of Expert Neural Processes},
    author={Qi Wang and Herke van Hoof},
    booktitle={Advances in Neural Information Processing Systems (2022NeurIPS)},
    editor={Alice H. Oh and Alekh Agarwal and Danielle Belgrave and Kyunghyun Cho},
    year={2022},
}

@inproceedings{ubukata2025graph,
  author={Ubukata, Toshihide and Zhang, Mingyue and Yamauchi, Takuto and Li, Jialong and Tei, Kenji},
  booktitle={2025 25th International Conference on Software Quality, Reliability, and Security Companion (QRS-C)}, 
  title={Graph-Contextual Reinforcement Learning for Efficient Exploration in Directed Controller Synthesis}, 
  year={2025},
  volume={},
  number={},
  pages={777-779},
  doi={10.1109/QRS-C65679.2025.00104}}

@inproceedings{SEAMS2026,
  title        = {Robust Exploration in Directed Controller Synthesis via Mixture-of-Experts Reinforcement Learning},
  author       = {Ubukata, Toshihide and Zhang, Mingyue and Wang, Zhiyao and Li, Nianyu and Li, Jialong and Tei, Kenji},
  booktitle    = {Proceedings of the 21st International Conference on Software Engineering for Adaptive and Self-Managing Systems (SEAMS)},
  year         = {2026},
  month        = {April}
}

\end{document}